\documentclass[11pt,a4paper]{article}
\usepackage[hyperref]{acl2021}
\usepackage{times}
\usepackage{latexsym}

\usepackage{microtype}

\aclfinalcopy 
\def\aclpaperid{2906} 

\usepackage{soul}
\usepackage{url}
\usepackage[utf8]{inputenc}
\usepackage{graphicx}
\usepackage{amsmath}
\usepackage{booktabs}
\usepackage{colortbl}
\usepackage{multirow}
\usepackage{setspace}  
\newcommand{\eat}[1]{}

\usepackage{quoting}

\newenvironment{ite}{                     
     \parskip 0cm \begin{itemize} \parskip 0cm \parsep 0cm \itemsep 0cm \topsep 0cm}{
        \end{itemize}} 
\newenvironment{enu}{                   
     \parskip 0cm \begin{list}{}{\parsep 0cm \itemsep 0cm \topsep 0cm}}{
       \end{list}} 
\newenvironment{des}{                 
     \parskip 0cm \begin{list}{}{\parsep 0cm \itemsep 0cm \topsep 0cm}}{
       \end{list}} 

\title{{P}roof{W}riter: {G}enerating Implications, Proofs, and Abductive Statements over Natural Language}

\author{
Oyvind Tafjord, Bhavana Dalvi Mishra, Peter Clark \\
Allen Institute for AI, Seattle, WA \\
\texttt{\{oyvindt,bhavanad,peterc\}@allenai.org} 
}

\eat{
\author{First Author \\
  Affiliation / Address line 1 \\
  Affiliation / Address line 2 \\
  Affiliation / Address line 3 \\
  \texttt{email@domain} \\\And
  Second Author \\
  Affiliation / Address line 1 \\
  Affiliation / Address line 2 \\
  Affiliation / Address line 3 \\
  \texttt{email@domain} \\}
}

\begin{document}
\maketitle
\begin{abstract}
Transformers have been shown to emulate logical deduction over natural language theories
(logical rules expressed in natural language), reliably assigning true/false labels
to candidate implications. However, their ability to {\it generate} implications of a theory
has not yet been demonstrated, and methods for reconstructing proofs of answers are imperfect.
In this work we show that a generative model, called ProofWriter, can reliably generate both
implications of a theory and the natural language proofs that support them. In particular, iterating a
1-step implication generator results in proofs that are highly reliable,
and represent actual model decisions (rather than post-hoc rationalizations).
On the RuleTaker dataset, the accuracy of ProofWriter's proofs exceed
previous methods by +9\% absolute, and in a way that 
generalizes to proof depths unseen in training and on out-of-domain problems.
We also show that generative techniques can perform a type of abduction with high precision:
Given a theory and an unprovable conclusion, identify a missing fact 
that allows the conclusion to be proved, along with a proof.
These results significantly improve the viability of neural methods for systematically reasoning over natural
language.\footnote{Datasets available at https://allenai.org/data/proofwriter}
\end{abstract}


\section{Introduction}
\begin{figure}[t]
\centering
     \includegraphics[width=0.8\columnwidth]{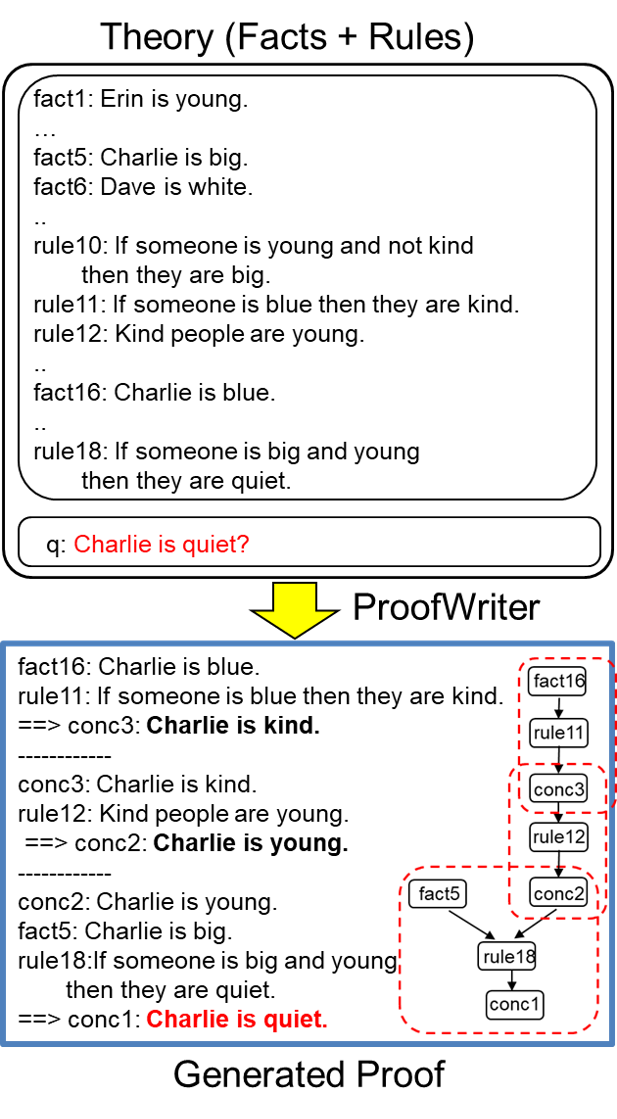}	   
     \vspace{-3mm}
\caption{Given facts, rules, and a question all expressed in natural language,
  ProofWriter answers the question and generates a proof of the answer. \label{proof}}
\vspace{-3mm}
\end{figure}

A fundamental goal for AI, dating back to its earliest years, is automated reasoning:
the ability to draw valid conclusions from explicitly provided knowledge
\cite{Mccarthy1959ProgramsWC}. However,
approaches relying on expressing knowledge in a formal representation language have
sometimes proved challenging \cite{musen1988brittleness}.
Recent work on RuleTaker \cite{ruletaker} demonstrated a modern approach to this goal, in which
transformers emulate deductive reasoning over statements expressed in {\it natural language},
by reliably assigning true/false labels to candidate implications. 
However, simply assigning true/false labels is limiting. For practical purposes,
systems should also generate proofs of those labels, so that their conclusions
can be verified and a human-understandable rationale be produced.
\eat{
ALTERNATIVE:
a reasoner should have additional capabilities: it should be able to
generate proofs of those labels so its conclusions can be verified and a
human-understandable rationale be produced; it shoud be able to
{\it generate} implications, not just assign true/false labels;
and it should be able to perform abduction to complete proofs
of unproven but known true facts. Our work here addresses these
three tasks.}

\begin{figure}[t]
\centering
     \includegraphics[width=0.85\columnwidth]{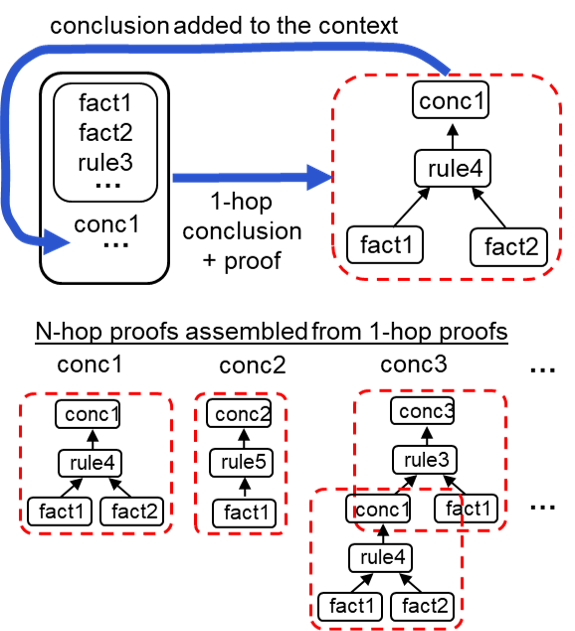}	   
\caption{ProofWriter iteratively generates 1-step implications and their
proofs, and adds implications back into into the context for deeper
reasoning. The stepwise proof fragments are assembled into full
proofs of N-hop conclusions. \label{architecture}}
\vspace{-3mm}
\end{figure}

Recent work on PRover, by \citet{prover}, provided first results towards this goal,
assembling proofs by first classifying which facts, rules, and connections
should be in the proof tree then using an Integer Linear Programming (ILP) module to enforce
consistency constraints. However,
the generated proofs were imperfect, and there were no guarantees that
the model ``believed'' the proofs that it was reciting, i.e., 
that its QA module would agree with the steps shown in the proof.
In this paper, we adopt a different approach, based on generation rather
than classification. Our system, ProofWriter, generates proofs
such as that shown in Figure~\ref{proof} by
iteratively generating 1-hop inferences and their (simple) proofs, adding
implications back into the context for deeper reasoning, and
assembling more complex proofs from the 1-hop fragments (Figure~\ref{architecture}).
As the accuracy of 1-hop inference is highly reliable, the accuracy
of deeper inference and their proofs is also high. This results in
proofs that substantially exceed the earlier method's accuracy, and also
reflect the model's internal decisions, rather than a post-hoc rationalization
(i.e., is a ``faithful'' proof \cite{Subramanian2020ObtainingFI}).

The generative approach also affords two other new capabilities.
First, ProofWriter generates implications that logically follow from a
NL (natural language) theory, allowing enumeration of consequences (rather than
only assigning truth values to pre-conjectured hypotheses).
Second, we demonstrate (a constrained form of) 
abduction: Given a theory and
an unprovable conclusion, identify a missing fact (if any) that allows the
conclusion to be proved when added to the theory, plus its proof.

We evaluate our work on a collection of natural language reasoning datasets,
including the RuleTaker datasets as well as several new variants.
We achieve state-of-the-art results in proof generation, and
strong new baselines for implication enumeration and abduction
over natural language theories. Our contributions
are thus:
\begin{enu}
\item[1.] A new method for proof generation for logical reasoning over natural language, that
obtains state-of-the-art results and is faithful to the model's internal decisions.
\item[2.] A method and baseline results for {\it generating} logical implications of statements in NL.
\item[3.] A method and baseline results for performing abduction over natural language statements.
\item[4.] New datasets to promote further research.
\end{enu}
These results significantly improve the viability
of neural methods for formal reasoning over language. 

\section{Related Work}



Our work builds on the RuleTaker line of research, in which
transformers learn to emulate a deductive reasoning {\it algorithm} \cite{ruletaker}.
Unlike other approaches to reasoning such as parsing to a formal language
\cite{semantic-parsing}, implementing a reasoning algorithm with neural components \cite{Weber2019NLPrologRW,Rocktschel2017EndtoendDP},
or SAT solving \cite{Selsam2018LearningAS}, these transformers emulate reasoning over language directly,
bypassing a formal representation.

PRover \cite{prover}, mentioned earlier, was the first system to also produce proofs in this context,
although its post hoc approach meant that proofs did not necessarily represent the actual model decisions.
\citet{Gontier2020MeasuringSG} also explored the generation of answers and proofs, but in the
context of rule {\it induction} with  ($\approx$ 10) fixed rules to induce. In contrast,
ProofWriter generates proofs from explicit NL rules (which may differ for each problem).
Similarly, formal theorem proving has explored proving mathematical theorems from
fixed, fundamental axioms, e.g., \cite{Polu2020GenerativeLM,Wang2020LearningTP},
while ProofWriter performs inference with differing sets of rules expressed in
natural language.


\begin{table*}
{\small
\begin{tabular}{|l|l|p{10.2cm}|} \hline
{\bf Proof (and Answer)} &  $CQ \to AP$ & 
Given theory $C$ and hypothesis fact $Q$, determine $Q$'s truth $A$ and proof $P$ (if any) \\
{\bf Enumeration} & $C \to I_{1},...,I_{n}$ & Given $C$, generate all implications $I_{i}$ that logically follow. \\
{\bf Abduction} & $CQ \to f_{m}$ & Given $C$ and an unprovable fact $Q$, identify a new fact $f_{m}$ that, when added to $C$, would make $Q$ true. \\ \hline
\end{tabular}
}
\vspace{-1mm}
\caption{The three tasks that ProofWriter performs. \label{tasks}}
\vspace{-3mm}
\end{table*}

Our work is also distinct from the large body of work on rationales and explanation.
Work on rationales aims to identify sentences (or phrases) that caused a model
to make a particular decision, but without an explanation of {\it why} that
rationale led to the answer (the model's reasoning is opaque), e.g.,
\cite{DeYoung2020ERASERAB,Narang2020WT5TT}.
  Similarly, work on explanations
  has sought to 
  generate human-style justifications, which again
  are typically supporting evidence rather than a fully-formed line of reasoning,
       and without explicit reasoning rules \cite{camburu2018snli,Jhamtani2020LearningTE,Inoue2020R4CAB}.
  In contrast, ProofWriter produces a deductive chain of reasoning
  from what is known to what is concluded, using a transformer retrained
  to reason systematically.


  \vspace{-1mm}
  
\section{Approach}


\vspace{-1mm}
\subsection{Definitions}

\vspace{-2mm}

\noindent
Let:
\vspace{-4mm}
\begin{ite}
\item $C$ be a {\it theory}, a set of English sentences $C$ consisting of facts $F$ and rules $R$, each expressing a logical fact or rule in English. (We also refer to $C$ as the {\it context}).
\item $Q$ be a {\it question}, a hypothesis fact in English whose truth is to be determined based solely on the information in $C$.
\item $A$ be an {\it answer}, where $A \in \{True,False\}$ (if reasoning using a closed-world assumption) or $A \in \{True,False,Unknown\}$ (open-world assumption).
\item $P$ be a {\it proof}, described shortly.
\item $I$ be an {\it implication}, a fact that logically follows from $C$.
\end{ite}
\vspace{-2mm}
\noindent
We define three tasks (also see Table~\ref{tasks}):
\begin{des}
\item[{\bf 1. proof (inc. QA):}] $CQ \to AP$: Given $C$ and hypothesis fact $Q$, what is the truth $A$ and proof $P$ (if any) of $Q$?
\item[{\bf 2. enumeration:}]  $C \to I_{1},...,I_{n}$: Which $I_{i}$ follow from $C$?
\item[{\bf 3. abduction}(restricted form)] $CQ \to f_{m}$: Which extra fact $f_{m}$ will make $Q$ true given $C$?
\end{des}


\noindent
We reuse (and add to) the RuleTaker datasets for our work, which include all five elements above.
An example of a RuleTaker theory (facts and rules), a query, and a proof generated by ProofWriter are shown in Figure~\ref{proof}.
Facts and rules are English statements, and implications 
are English statements that logically follow from those facts and rules.
The original datasets were generated from synthetic logic programs and their implications,
using natural language patterns to produce the English forms.


\subsection{Semantics}

Following prior work, we adopt the semantics of Datalog \cite{datalog}: A fact is true if it is either known (i.e., explicitly
stated in the context $C$), or (recursively) is the conclusion of a rule whose conditions are true (is ``supported''). For handling negation,
we use two alternative Datalog semantics: The first, following prior work, makes the closed-world assumption (CWA) and uses negation as
failure (NAF), so that any fact not provable is assumed false. Under this semantics, negated facts and
 negative rule conclusions are not allowed (redundant under the CWA). The second makes an open-world
assumption (OWA), and does allow negative facts and rule conclusions. Under this semantics, a third
truth value {\it Unknown} is also possible.

\subsection{Proof Representation}

We define a proof $P$ of a fact $f_q$ as a directed acyclic graph $(N,E)$ with nodes $n \in N$ and (directed, untyped) edges $e \in E$.
Each node in $P$ is either a fact $f$ (a ground literal) or a rule $r$ (a logical implication), expressed in English.
Each edge in the proof either connects a fact to a rule, denoting that the fact helps satisfy the rule's
condition, or connects a rule to a fact, denoting that the fact follows from the instantiated rule.
Thus nodes in any branch of the proof will alternate between facts and rules. Note this definition
differs from (and is richer than) that in PRover, where intermediate conclusions were not part of the proof.

\begin{figure}
 \centering
     \includegraphics[width=0.8\columnwidth]{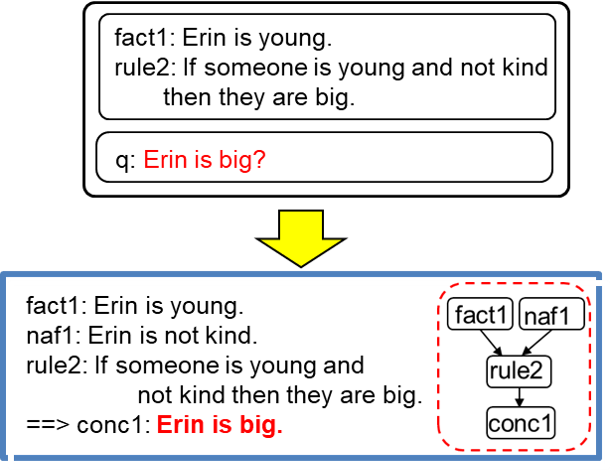}	   
    \vspace{-1mm}
    \caption{An example proof that includes a negated (negation-as-failure) fact.  \label{naf-proof}}
     \vspace{-3mm}
 \end{figure}

Facts in the proof are one of three types: known facts $f_{i} \in F$, negated facts $f_{naf}$ that cannot be proven (false under negation-as-failure (NAF)),
and facts $f_{conc}$ that are the conclusions of rules. 
$f_{i}$ and $f_{naf}$ are leaf nodes of the proof, while the $f_{conc}$ are intermediate nodes within the proof.
Note that $f_{naf}$ and $f_{conc}$ are by definition not in $F$. 
Example proofs are shown in Figures~\ref{proof} and~\ref{naf-proof}.

\subsection{Proof Encoding \label{proof-encoding}}

As we wish to generate proofs, we need to encode $P$ as a linear structure that can be output by a generative model.
Facts and rules in the context are explicitly labeled with identifiers (fact1, ..., rule1, ...) that
the proof can refer to, see Figures~\ref{proof} and~\ref{naf-proof}.\footnote{
In practice we name these sent1, sent2, ... without a fact/rule distinction, but for expository purposes it is helpful to use different identifiers.}
Then, in the linear proof, rule nodes are denoted by their identifier (rule1, ...), while
fact nodes are denoted by three types of identifiers: fact1, fact2, ... for facts in the context;
naf1, naf2, ... for facts {\it not} in the context and assumed false; and conc1, conc2, ... for facts
concluded by rules. To decode the naf* and conc* identifiers (which by definition are not in the context),
an additional sequence of the form ``with conc1: {\it sentence1}. conc2: {\it sentence2}. ...'' is
appended to the proof.

To linearize the proof in a format convenient for a generative model, we conjoin rules and their conclusions using a ``\%'' symbol, express conjunctive rule
conditions with a ``\&'' symbol, and use ``\#'' to denote the inverse implication (``$\leftarrow$''). We then express the tree using Polish notation.
E.g., the proof tree ``((fact1 \& fact2) $\rightarrow$ rule1 $\rightarrow$ conc1)'' (i.e., fact1 and fact2 satisfy rule1, concluding conc1)
 would be expressed ``\# rule1\%conc1 \& fact1 fact2''.
Thus the 3-step proof from Figure~\ref{proof} is encoded:
\begin{quote}
\# rule18\%conc1 \& fact5 \# rule12\%conc2 \# rule11\%conc3 fact16 ; with conc1: Charlie is quiet. ; conc2: Charlie is young. ; conc3: Charlie is kind.
\end{quote}
If the question is a known fact, the ``depth 0 proof'' is simply the fact itself (e.g., fact1). If no proof exists, the symbol ``None'' is used.

\subsection{Models \label{model}}

The ProofWriter models are built on top of the text-to-text pretrained T5 transformer \cite{Raffel2020ExploringTL} (T5-11B). We use different textual prompts for the different tasks.
For the task of generating an answer and a proof, the input to the model is of the form:
`` \$question\$ = {\it question} ; \$context\$ = {\it theory-sentences}'', 
for example:
``\$question\$ = Erin is big. ; \$context\$ = sent1: Erin is young. sent2: If ...''
The output is of the form:
``\$answer\$ = {\it True/False/Unknown} : \$proof\$ = {\it proof} ;'', 
where {\it proof} is encoded as described in Section~\ref{proof-encoding}.
For training instances where multiple outputs are valid, we select a single one at random (for multiple proofs, we select among the shortest proofs).
Appendix~\ref{appendix-inputs-outputs} lists the hyperperameters and gives input/output examples for each task. 


\subsection{Task 1: Proof Generation \label{two-models}}

We evaluate two methods of proof generation:
\begin{des}
\item[{\bf All-At-Once:}] We train a model to generate the full proof and answer in one go (theory + question $\rightarrow$ answer + proof).
\item[{\bf Iterative:}] We first train a model to generate a single 1-step implication (theory $\rightarrow$ implication + 1-step-proof),
where the implication follows from a single rule application.
Then at test time, we apply this model iteratively,
adding each implication to the theory and repeating until no more implications can be found (i.e., exhaustive forward-chaining).
The proof for any given implication can then be assembled from the 1-step-proof fragments (Figure~\ref{architecture}).
\end{des}

\subsubsection{All-At-Once ProofWriter (``All'')}

The All-At-Once model is trained directly on $CQ \rightarrow AP$ examples in the datasets ($P$ = ``None'' if there is no proof of $Q$).
Section~\ref{model} describes the i/o format, and Appendix~\ref{appendix-all-at-once-input-output} shows an example.

\subsubsection{Iterative ProofWriter (``Iter'') \label{train-iterative}}

{\bf Training:} To train the Iterative model, for each theory $C$ in the training data,
we create an augmented set of training examples with one sequence of iteratively inferred facts
in turn, each using $C$ plus the previously inferred facts.
For example, if theory $C_{1}$ implies $I_{1}$, $I_{2}$, and $I_{3}$, then we create four
training examples
$C_{1} \rightarrow I_{1}$,
$C_{1} \cup \{I_{1}\} \rightarrow I_{2}$,
$C_{1} \cup \{I_{1},I_{2}\} \rightarrow I_{3}$, and
$C_{1} \cup \{I_{1},I_{2},I_{3}\} \rightarrow$ ``None''.
The order of adding the $I_{i}$ is random but constrained such that if a later implication depends
on an earlier one, the earlier one must be inferred first. For example,
if the proof of $I_{3}$ depends on $I_{2}$ (determined by inspecting the gold proofs),
$I_{2}$ must be in the context before $I_{3}$ is inferred. This ensures that all
example inferences are depth 1 (i.e., a single rule application). An example input/output for one step is shown in Appendix~\ref{appendix-iterative-input-output}.

{\bf Testing:} To answer and provide the proof for a particular question/implication,
the model generates all implications and their proofs by iteratively applying
the model until no more implications (the implication ``None'') is generated.
It then looks for the question among them.
If found, the answer is {\it True} with the
proof given. The model also looks for the negation of the question\footnote{To negate
a question, a model can be trained for this straightforward task. Here, as our
question language is simple, a simple regex to add/remove a ``not'' suffices.}
and its proof. If found, the answer is {\it False} with the proof given.
Otherwise, there is no proof (proof = ``None'') and the answer is  {\it False} (for positive questions, CWA),  {\it True} (for negative questions, CWA), or  {\it Unknown} (any question, OWA).

\subsection{Task 2: Implication Enumeration \label{enumeration-method}}

A second desirable reasoning skill is {\it enumerating} implications of a theory (rather than just
assign {\it True/False} to a hypothesis). This capability is important for practical application of the technology.
In fact, the Iterative ProofWriter already does this by design, a substantial advantage.
To evaluate this (later), we compare this with an ``all at once'' strategy of generating
{\it all} implications as a single output string, analogous to the All-At-Once strategy for generating
the full proof as a single string. For training this All-At-Once enumerator, and testing both,
we gather the list of {\it all} implications $I_{i}$ of each theory $C$ in the train/test data.
Each train/test example is of then of the form: given $C$, predict all the $I_{i}$. An example input/output is in Appendix~\ref{appendix-enumeration-input-output}.


\subsection{Task 3: Abduction (Single Fact) \label{abduction-method}}

A third desirable reasoning skill is {\it abduction} over natural language theories, again
made possible by generative models. Abduction has previously been studied extensively
in formal logic, e.g., \cite{Konolige1997AbductiveTI}, and in
NLP, e.g., \cite{Hobbs1993InterpretationAA,Bhagavatula2020AbductiveCR}.
Here we evaluate whether a generative approach can combine logic and NLP, performing logical
abduction over natural language knowledge.
We do this for a restricted
form of abduction, namely single-fact abduction: Given a theory $C$
and a possible implication $Q$ not provable from $C$, identify a new fact $f_{m}$ (other than the trivial $Q$ itself) such that
$C \cup \{f_{m}\}$ implies $Q$.

We restrict this task to the OWA (open-world) setting where questions can naturally have unknown truth values. To train and test an abductive model over our datasets, we create an abductive version
as follows: For each theory $C$ in the train/test data, for each unprovable fact $Q$,
identify all alternative ``missing facts'' $factM$ that, when added to $C$, make $Q$ True.
To do this, recall that each NL theory was originally generated from a formal one
$C_{formal}$ in a formal representation language (Datalog). 
We first exhaustively enumerate all possible $Q_{formal}$ and $factM_{formal}$ in the
formal language (this is feasible as the space of predicates and individuals is small),
then use a theorem prover to test if $C_{formal} \cup \{factM_{formal}\}$ implies $Q_{formal}$
for all pairs $(factM_{formal},Q_{formal})$.
For each success, we generate the NL equivalents $Q$ and $factM$ using simple NL generation templates.
We then collect the alternative $factM$s for each $Q$. The abduction task is then,
given $C$ and $Q$, identify the set of all alternative $factM$s,
i.e.:
\vspace{1mm} \\
\hspace*{1cm} $C, Q \rightarrow factM_{1}, ..., factM_{i}$
\vspace{1mm} \\
If there is no single $factM$ that can be added to make $Q$ true, then
the
symbol ``None'' is output.

\section{Datasets}

We now evaluate ProofWriter on these three tasks.
We use the original RuleTaker D* datasets \cite{ruletaker}, plus we create two new variants:
The first (CWA) is similar to the original except it fixes some minor inconsistencies concerning negation (details in Appendix~\ref{dataset-repairs}).
The second (OWA) is also similar to the original, except reasoning uses an open-world assumption.

We denote these as D*(orig), D*(CWA), and D*(OWA).
Each example in each dataset contains a theory $C$, a question $Q$, the answer $A$ ({\it True/False/Unknown}), and all
possible proofs $P_{1},...,P_{n}$ for that answer (if provable).\footnote{
The domain is small enough that all proofs can be enumerated. However, there still can be a large number, e.g., some
D5 questions have over 3000 possible proofs.} Each theory is also accompanied with all possible proofs of all possible
implications, as auxiliary annotations.

The D* datasets comprise five datasets, named {\bf D0, D1, D2, D3, D5}, each containing 100k questions. 
In each dataset, theories and questions are expressed in templated English
(e.g., Figure~\ref{proof}), questions can be positive or negated facts (e.g., ``Charlie is not quiet?''), and answers are equally divided
into {\it True/False} (and {\it Unknown}, for the OWA versions). Each dataset contains questions whose answers require reasoning up to depths $D$ ($D$ = 0, 1, 2, 3, 5).
Thus, for example, all questions in D0 are lookup questions, requiring no inference. Each dataset is split 70/10/20 into train/dev/test.

To test generalization, we also use two other datasets from the original RuleTaker work:

\noindent
{\bf Birds-Electricity:} These 6 test-only datasets use small, real-world theories written by hand (one per dataset)
to test out-of-distribution model performance. Details are in Appendix~\ref{dataset-details-birds}.

\noindent
{\bf ParaRules:} This dataset contains 40k questions against 2k theories expressed in paraphrased natural
language, obtained through crowdsourcing. This dataset tests transfer to more natural expressions of knowledge.
Details are in Appendix~\ref{dataset-details-para}.

\section{Experiments and Results}


\subsection{Task 1: Proof Generation (Comparison with Prior Work)}

First, we compare ProofWriter's ability to generate proofs with PRover, the current state-of-the-art.
We evaluate both answer accuracy and proof correctness. For proof correctness, for a fair comparison, we
ignore the intermediate conclusion nodes (which PRover does not generate).
We then use the same strict scoring metric as in PRover (called FA or Full Accuracy in the PRover paper):
the proof graph must {\it exactly} match a gold proof (i.e., be perfectly correct); otherwise, the
proof scores 0.

\subsubsection{Generating Answers and Proofs}

\begin{table}
\centering
{\small
\setlength{\tabcolsep}{3pt}	
\begin{tabular}{lcccccc}
\hline
 &  & \multicolumn{2}{c}{\bf{Answer}} & \multicolumn{2}{c}{\bf{Proof}}\\
Depth & \# qns & PRover & ProofWriter & PRover & ProofWriter\\
\hline
0 & 6299 & \cellcolor{blue!50} 100 & \cellcolor{blue!50} 100 & \cellcolor{blue!49} 98.4 & \cellcolor{blue!50} 99.6\\
1 & 4434 & \cellcolor{blue!50} 99.0 & \cellcolor{blue!50} 99.1 & \cellcolor{blue!47} 93.1 & \cellcolor{blue!49} 98.7\\
2 & 2915 & \cellcolor{blue!49} 98.8 & \cellcolor{blue!49} 98.6 & \cellcolor{blue!42} 84.8 & \cellcolor{blue!49} 97.3\\
3 & 2396 & \cellcolor{blue!50} 99.1 & \cellcolor{blue!49} 98.5 & \cellcolor{blue!40} 80.5 & \cellcolor{blue!47} 94.4\\
4 & 2134 & \cellcolor{blue!49} 98.8 & \cellcolor{blue!49} 98.7 & \cellcolor{blue!36} 72.4 & \cellcolor{blue!46} 91.0\\
5 & 2003 & \cellcolor{blue!50} 99.3 & \cellcolor{blue!50} 99.3 & \cellcolor{blue!33} 65.1 & \cellcolor{blue!43} 86.4\\\hline
All & 20192 & \cellcolor{blue!50} 99.3 & \cellcolor{blue!50} 99.2 & \cellcolor{blue!44} 87.1 & \cellcolor{blue!48} 96.2\\
\hline
\end{tabular}
}
\caption{[Task 1: Proof Generation] 
  Systems trained and tested on D5(orig), 
  showing the breakdown
by depth of proof required to answer each question.
ProofWriter generates significantly more correct proofs for all depths, achieving a new SOTA on this task.
\label{basic-results}}
\vspace{-3mm}
\end{table}

We use the same IID (independent, identically distributed) data used for PRover (train/test on dataset D5(orig)).
The results are in Table~\ref{basic-results}, showing accuracies for questions requiring increasingly deeper depths
of reasoning to answer.
The ProofWriter's results are for the All-At-Once model. (The Iterative model
scores are almost identical, see later Table~\ref{comparison-iid}.) While answer accuracy is almost perfect for both systems,
ProofWriter generates {\bf substantially more correct proofs} (last line, +9\% absolute), and without the complexity
of PRover's heuristic assembly of proof graphs using ILP.

\subsubsection{Performance on OOD Rulesets \label{proofs-on-ood}}

\eat{SUBMISSION
  We also compared ProofWriter's and PRover's ability to generalize to the
hand-authored Birds-Electricity rulesets, zero shot. We find that
ProofWriter {\bf generate more correct proofs} than PRover in
this out-of-domain (OOD) setting (84.5\% for the All-At-Once ProofWriter,
97\% for the Iterative ProofWriter, vs. 80.5\% PRover). Details are
in Appendix~\ref{proofs-on-ood-details}.
We also find ProofWriter obtains more correct proofs (+3\%) than PRover on
the ParaRules dataset. Details are in Appendix~\ref{appendix-pararules}.
}

We compared ProofWriter's and PRover's ability to generalize to the hand-authored Birds-Electricity
rulesets, zero shot. 
These rulesets are out-of-domain (OOD), as their English
is not templated and is stylistically different to the training data.
We compare the PRover and All-At-Once (``All'') ProofWriter models trained on D5,
plus the Iterative ProofWriter (``Iter'') trained on D0-D3 theories.
The models do not see any Birds-Electricity examples during training.
The results in Table~\ref{birds} show that ProofWriter's proof generation
transfers well zero-shot to these hand-authored datasets, with
84.5\% proof correctness for All-At-Once, and 97\% for the Iterative ProofWriter,
indicating better out-of-domain generalization for the Iterative version.
Both ProofWriter models significantly outperform PRover (80.5\%).

We also find ProofWriter obtains more correct proofs (+3\%) than PRover on
the ParaRules dataset. Details are in Appendix~\ref{appendix-pararules}.

\begin{table}
\centering
{\small
\setlength{\tabcolsep}{3pt}	
\begin{tabular}{lcccccccc}
\hline
 &  & \multicolumn{3}{c}{\bf{Answer}} & \multicolumn{3}{c}{\bf{Proof}}\\
 &  & \multirow{2}{*}{PRover} & \multicolumn{2}{c}{ProofWriter} & \multirow{2}{*}{PRover} & \multicolumn{2}{c}{ProofWriter}\\
Test & \# qns & &  All & Iter &  & All & Iter\\
\hline
Birds1 & 40 & \cellcolor{blue!48} 95.0 & \cellcolor{blue!50} 100 & \cellcolor{blue!48} 95.0 & \cellcolor{blue!46} 92.5 & \cellcolor{blue!50} 100 & \cellcolor{blue!48} 95.0\\
Birds2 & 40 & \cellcolor{blue!48} 95.0 & \cellcolor{blue!50} 100 & \cellcolor{blue!48} 95.0 & \cellcolor{blue!48} 95.0 & \cellcolor{blue!50} 100 & \cellcolor{blue!48} 95.0\\
Elec1 & 162 & \cellcolor{blue!50} 100 & \cellcolor{blue!48} 96.9 & \cellcolor{blue!50} 100 & \cellcolor{blue!48} 95.1 & \cellcolor{blue!48} 96.9 & \cellcolor{blue!50} 100\\
Elec2 & 180 & \cellcolor{blue!50} 100 & \cellcolor{blue!49} 98.9 & \cellcolor{blue!50} 100 & \cellcolor{blue!46} 91.7 & \cellcolor{blue!49} 98.9 & \cellcolor{blue!50} 100\\
Elec3 & 624 & \cellcolor{blue!45} 89.7 & \cellcolor{blue!46} 92.0 & \cellcolor{blue!48} 95.5 & \cellcolor{blue!36} 71.8 & \cellcolor{blue!46} 92.0 & \cellcolor{blue!48} 95.5\\
Elec4 & 4224 & \cellcolor{blue!42} 84.8 & \cellcolor{blue!42} 83.3 & \cellcolor{blue!49} 97.1 & \cellcolor{blue!40} 80.6 & \cellcolor{blue!41} 82.0 & \cellcolor{blue!49} 97.1\\\hline
All & 5270 & \cellcolor{blue!43} 86.5 & \cellcolor{blue!43} 85.5 & \cellcolor{blue!48} 97.0 & \cellcolor{blue!40} 80.5 & \cellcolor{blue!42} 84.5 & \cellcolor{blue!48} 97.0\\
\hline
\end{tabular}
}
\caption{[Task 1: Proof Generation] 
  Training on D5, test on Birds-Electricity. Both ProofWriter versions (``{\bf All}'' for All-At-Once, ``{\bf Iter}'' for Iterative)
  outperform PRover overall in both answer and proof correctness. The Iterative model 
  is also
significantly more robust. \label{birds}}
\vspace{-3mm}
\end{table}

\subsection{Task 1: Proof Generation (All-At-Once vs. Iterative)}

Second, we compare our two approaches to proof generation, All-At-Once vs.
Iterative, in more detail. We show that although they have 
almost identical performance for proofs with depths seen in training,
{\bf the Iterative model generalizes better to proofs of longer depths} than
seen in training.
For these comparisons, we use the new D*(CWA) datasets (which fix some minor
errors in D*(orig)), and also the D*(OWA) datasets to explore performance
in an open-world setting.

\eat{
\subsubsection{Training the Iterative Model}
We train the iterative model as described earlier (Section~\ref{train-iterative}),
using the ($\sim$ 5k) theories from the D3 dataset (train). We also add
some ($\sim$ 2k each) D0-D2 theories, to have more examples of theories with
less conclusions. We include this iterative dataset in our dataset release
(both a CWA and OWA version).
}

\subsubsection{Comparison (IID Test Set)}

We train the All-At-Once model on D5 (train), and the Iterative model
using the method described in Section~\ref{train-iterative},
using the ($\sim$ 5k) theories from D3 (train) plus $\sim$ 20\% of
the D0-D2 (train) theories.\footnote{We include D0-D2 theories 
to have more examples of theories with fewer conclusions.
The derivative iterative training data is included in our dataset
release.} We then test both models on D5 (test).
We measure both answer and proof accuracies, and also break down the
results by proof depth (using ``N/A'' as the proof depth for questions
that are not provable).
The D5 test set has 2k questions at each proof depth, plus 8k unprovable questions (proof = ``None'', depth = ``N/A').\footnote{Note this breakdown is slightly different from the one in Table~\ref{basic-results} where the depth used the original RuleTaker annotations which included a depth for questions without proofs, based on the deepest proof search that fails. We retained that convention in Table~\ref{basic-results} for best comparison with PRover.}

The results are shown in Table~\ref{comparison-iid}, and show
that {\bf both ProofWriter versions have similar, high proof correctness (95\%+)}
on the test set, even though some proofs are highly complex. 

\begin{table}
\centering
{\small
\setlength{\tabcolsep}{3pt}	
\begin{tabular}{lccccccccc}
\hline
 & \multicolumn{4}{c}{\bf{Answer}} & \multicolumn{4}{c}{\bf{Proof}}\\
 & \multicolumn{2}{c}{CWA} & \multicolumn{2}{c}{OWA} & \multicolumn{2}{c}{CWA} & \multicolumn{2}{c}{OWA}\\
Depth & All & Iter & All & Iter & All & Iter & All & Iter\\
\hline
N/A & \cellcolor{blue!50} 99.0 & \cellcolor{blue!50} 99.7 & \cellcolor{blue!50} 99.4 & \cellcolor{blue!50} 99.9 & \cellcolor{blue!50} 99.0 & \cellcolor{blue!50} 99.7 & \cellcolor{blue!50} 99.4 & \cellcolor{blue!50} 99.9\\
0 & \cellcolor{blue!50} 100 & \cellcolor{blue!50} 100 & \cellcolor{blue!50} 100 & \cellcolor{blue!50} 100 & \cellcolor{blue!50} 100 & \cellcolor{blue!50} 100 & \cellcolor{blue!50} 100 & \cellcolor{blue!50} 100\\
1 & \cellcolor{blue!50} 99.9 & \cellcolor{blue!50} 99.8 & \cellcolor{blue!50} 100 & \cellcolor{blue!50} 99.3 & \cellcolor{blue!50} 99.6 & \cellcolor{blue!48} 95.4 & \cellcolor{blue!50} 99.7 & \cellcolor{blue!49} 97.8\\
2 & \cellcolor{blue!50} 99.9 & \cellcolor{blue!50} 99.5 & \cellcolor{blue!50} 99.9 & \cellcolor{blue!50} 99.7 & \cellcolor{blue!49} 98.3 & \cellcolor{blue!46} 91.7 & \cellcolor{blue!49} 98.6 & \cellcolor{blue!49} 97.3\\
3 & \cellcolor{blue!50} 100 & \cellcolor{blue!50} 99.7 & \cellcolor{blue!50} 100 & \cellcolor{blue!50} 99.2 & \cellcolor{blue!48} 95.8 & \cellcolor{blue!45} 90.4 & \cellcolor{blue!48} 96.9 & \cellcolor{blue!49} 97.1\\
4 & \cellcolor{blue!50} 100 & \cellcolor{blue!50} 99.7 & \cellcolor{blue!50} 99.9 & \cellcolor{blue!50} 99.1 & \cellcolor{blue!47} 93.1 & \cellcolor{blue!44} 88.9 & \cellcolor{blue!47} 94.8 & \cellcolor{blue!48} 96.5\\
5 & \cellcolor{blue!50} 99.9 & \cellcolor{blue!49} 98.9 & \cellcolor{blue!50} 100 & \cellcolor{blue!49} 98.8 & \cellcolor{blue!45} 89.3 & \cellcolor{blue!44} 87.8 & \cellcolor{blue!46} 91.4 & \cellcolor{blue!43} 86.4\\\hline
All & \cellcolor{blue!50} 99.6 & \cellcolor{blue!50} 99.7 & \cellcolor{blue!50} 99.7 & \cellcolor{blue!50} 99.6 & \cellcolor{blue!49} 97.2 & \cellcolor{blue!48} 95.4 & \cellcolor{blue!49} 98.0 & \cellcolor{blue!49} 97.6\\
\hline
\end{tabular}
}
\caption{[Task 1] Comparison of All-At-Once (``{\bf All}'') vs. Iterative (``{\bf Iter}'') ProofWriter models, trained on D5 and D0-D3 respectively, and tested on D5. \label{comparison-iid}}
\vspace{-3mm}
\end{table}

\subsubsection{Generalization to Unseen Depths}

We also wish to see how well the models can generate proofs at
depths unseen during training. To do this, we train an All-At-Once model on D3,
and use the same Iterative model as earlier (trained on iterative examples
from theories up to depth 3). We test on D5.
As D5 contains problems at greater depths than those seen during
training, we can observe the models' ability to generalize.
We compare with both the CWA and OWA versions of our datasets.

The results are shown in Table~\ref{comparison-ood}.
As can be seen, the
All-At-Once model has quite poor generalization for generating longer proofs
than seen in training, while {\bf the Iterative model is more robust} (red box).

\begin{table}
\centering
\includegraphics[width=1\columnwidth]{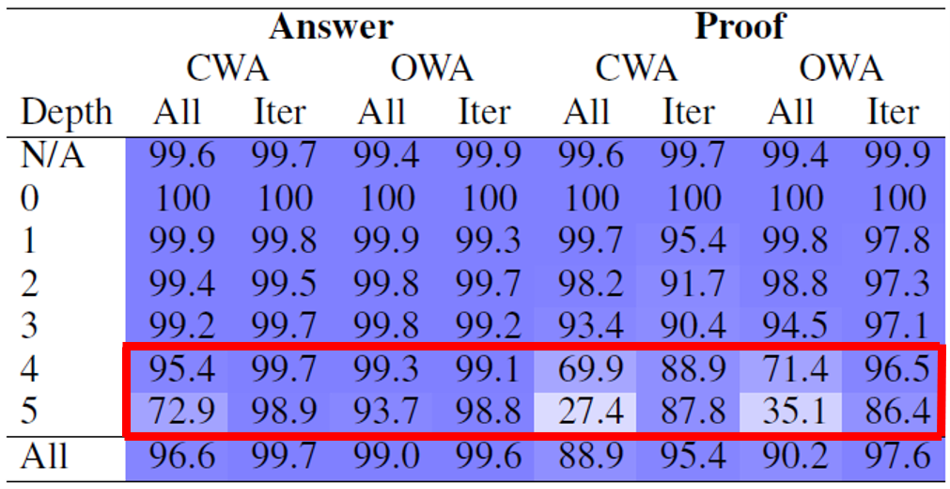}
\vspace{-5mm}
\caption{[Task 1] 
Comparison of the All-At-Once vs. Iterative ProofWriter models,
trained on D3 and tested on D5.
While scores are mostly similar throughout, the iterative model generalizes
substantially better to generate proofs of depths unseen during training (red box).
\label{comparison-ood}}
\vspace{-3mm}
\end{table}

\begin{figure}[t]
  \vspace{-2mm}
\centering
     \includegraphics[width=0.85\columnwidth]{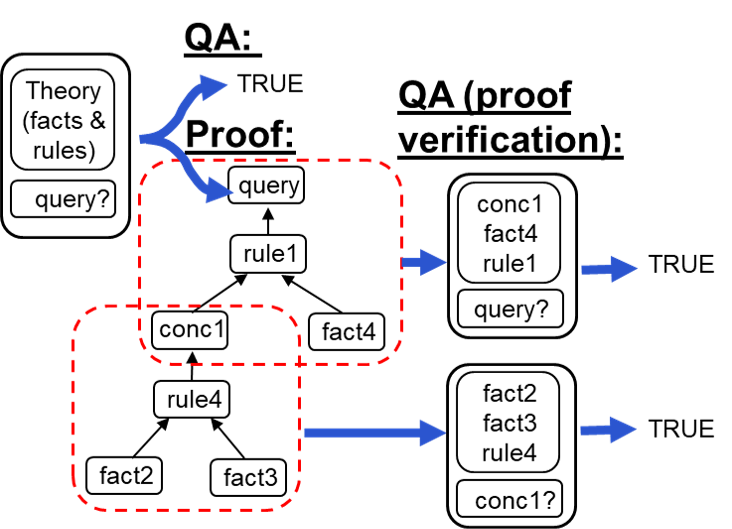}	   
\caption{[Task 1] All-At-Once proofs can be verified by checking each step as a separate QA query. \label{verification-fig}}
\vspace{-3mm}
\end{figure}

\subsection{Verifying All-At-Once Proofs \label{verification}}

Proofs from the Iterative ProofWriter have an additional desirable property:
each proof step is one that the model explicitly took during the iteration,
i.e., the model ``believes'' the step. In contrast, the All-At-Once proofs are
a post hoc generated string of symbols, and may not reflect steps that ProofWriter
would actually make. However, because proofs include intermediate conclusions,
we can alleviate this concern by {\it verifying} individual steps in
the All-At-Once proofs. For example, if a generated proof step states that fact2 + fact3 + rule4
implies conc1, we can simply ask ProofWriter in QA mode if this is true
(Figure~\ref{verification-fig}).
Given the almost perfect performance for such simple depth 1 questions
in QA mode (with no distractor facts or rules), the ability to verify a correct proof corresponds to the accuracy of
correctly generating the correct intermediate conclusions conc* in the
first place. (Note that an unverified proof is not necessarily wrong,
rather cannot be verified as right).
OWA proofs can be fully verified in this way. 
For CWA theories with NAFs, the verification is
only partial as NAFs are presumed negative statements which require the full theory to verify.

We measured the percentage of correct, verified proofs, shown in Table~\ref{intermediates}.
Provided proofs are within the depths seen during training, almost
all correct proofs can be verified. However, at depths deeper than seen at
training, the proportion that can be verified drops rapidly.
In contrast, the Iterative ProofWriter's proofs are always verified,
as by definition they are assembled from single step inferences that
the model actually took.

\begin{table}
\centering
{\small
\setlength{\tabcolsep}{3pt}	
\begin{tabular}{lccccc}
\hline
 & \multicolumn{4}{c}{\bf Verified Proofs} \\
 & \multicolumn{2}{c}{CWA} & \multicolumn{2}{c}{OWA}\\
 & \multicolumn{2}{c}{Train on:} & \multicolumn{2}{c}{Train on:}\\
Depth & D3 & D5 & D3 & D5\\
\hline
N/A & \cellcolor{blue!50} 99.6 & \cellcolor{blue!50} 99.0 & \cellcolor{blue!50} 99.4 & \cellcolor{blue!50} 99.4\\
0 & \cellcolor{blue!50} 100 & \cellcolor{blue!50} 100 & \cellcolor{blue!50} 100 & \cellcolor{blue!50} 100\\
1 & \cellcolor{blue!50} 99.8 & \cellcolor{blue!50} 99.6 & \cellcolor{blue!50} 99.7 & \cellcolor{blue!50} 99.7\\
2 & \cellcolor{blue!49} 98.2 & \cellcolor{blue!49} 98.3 & \cellcolor{blue!49} 98.6 & \cellcolor{blue!49} 98.6\\
3 & \cellcolor{blue!47} 93.2 & \cellcolor{blue!48} 95.8 & \cellcolor{blue!47} 94.3 & \cellcolor{blue!48} 96.8\\
4 & \cellcolor{blue!33} 66.7 & \cellcolor{blue!46} 92.9 & \cellcolor{blue!33} 66.1 & \cellcolor{blue!47} 94.6\\
5 & \cellcolor{blue!7} 13.8 & \cellcolor{blue!45} 89.3 & \cellcolor{blue!8} 16.4 & \cellcolor{blue!45} 90.8\\\hline
All & \cellcolor{blue!44} 87.2 & \cellcolor{blue!49} 97.2 & \cellcolor{blue!44} 87.9 & \cellcolor{blue!49} 97.9\\ \hline
\end{tabular}
}
\caption{[Task 1: Proof Generation] The All-At-Once model's ability to verify its proofs.
For proofs within depths seen during training,
almost all correct proofs (Tables~\ref{comparison-ood} and \ref{comparison-iid}, columns 5 and 7)
can be verified. However, for proofs at unseen depths, the proportion that
can be verified drops rapidly (trained on D3, test on depths 4,5).
In contrast, Iterative ProofWriter's proofs are always verified,
by definition of its algorithm.
\label{intermediates}}
\vspace{-3mm}
\end{table}

\eat{SUBMISSION
  A disadvantage of All-At-Once proofs is that they may not reflect actual
model decisions (as they are a post hoc generated string of symbols).
We attempted to alleviate this by {\it verifying} each individual step
using ProofWriter in QA mode. We found that {\it if} the
proofs were within the same depth used in training, almost all
All-At-Once proofs were verified as correct, but that verified correctness
dropped significantly at higher depths, a limitation for the All-At-Once
approach (details in Appendix~\ref{verification-details}).
}


\subsection{Task 2: Implication Enumeration}

Third, we evaluate ProofWriter's performance on a new task, namely {\it enumerating} implications
of a theory (rather than just assign {\it True/False} to a hypothesis). We compare
the All-At-Once and Iterative strategies as described in Section~\ref{enumeration-method}.

To train All-At-Once, and test both, we created an enumerative dataset of $C \rightarrow \{I_{1},...,I_{n}\}$
examples (Section~\ref{enumeration-method}).
For this we sample theories $C$ in the D0-D3 datasets and gather the list of all implications $I_{i}$ for each theory $C$.
We call this enumerative dataset {\bf D3+Enum}. 
We similarly create a {\bf D5-Enum} dataset from theories in (only) D5 to test OOD conclusion generation.
We create CWA and OWA versions of both.

We train All-At-Once on D3-Enum (train), then test both models on D3-Enum (test) and D5-Enum (test).
For metrics, we measure F1 scores by comparing the individual predicted implications with the gold $I_{i}$,
as well as the exact-match correctness of the predicted {\it set} of implications $\{I_{1},...,I_{n}\}$
(one point if the set {\it exactly} matches the gold, bar ordering, zero otherwise).
The results are shown in Table~\ref{enumeration}, and show that the {\bf Iterative ProofWriter is
better at implication enumeration} than the simple All-At-Once strategy. In particular,
the All-At-Once strategy struggles for problems at depths unseen in training (second row),
although it does well on its own test set despite the complicated unordered output it has to generate (up to 16 different implications in D3, 21 in D5).

\begin{table}
\centering
{\small
\setlength{\tabcolsep}{3pt}	
\eat{
\begin{tabular}{|ll|cc||cc|} \hline
        &	& \multicolumn{2}{|c||}{F1} & \multicolumn{2}{||c|}{Accuracy} \\
Test	& Count & All   & Iter	& All	& Iter \\	 \hline
D3+Enum & 3135  &     99.36  &  99.57  &      95.47   &   99.04 \\ 
D5-Enum & 948   &     94.76  &  99.43  &      48.95   &   94.83 \\ \hline
\end{tabular}
}
\begin{tabular}{lccccccccc}
\hline
 & \multicolumn{4}{c}{\bf{F1}} & \multicolumn{4}{c}{\bf{Accuracy}}\\
Enum & \multicolumn{2}{c}{CWA} & \multicolumn{2}{c}{OWA} & \multicolumn{2}{c}{CWA} & \multicolumn{2}{c}{OWA}\\
Test  & All & Iter & All & Iter & All & Iter & All & Iter\\
\hline
D3+  & \cellcolor{blue!49} 98.9 & \cellcolor{blue!50} 99.8 & \cellcolor{blue!50} 99.4 & \cellcolor{blue!50} 99.6 & \cellcolor{blue!46} 92.5 & \cellcolor{blue!49} 98.8 & \cellcolor{blue!48} 95.5 & \cellcolor{blue!50} 99.0\\
D5 & \cellcolor{blue!47} 94.5 & \cellcolor{blue!50} 99.5 & \cellcolor{blue!47} 94.8 & \cellcolor{blue!50} 99.4 & \cellcolor{blue!22} 44.6 & \cellcolor{blue!47} 93.9 & \cellcolor{blue!24} 48.9 & \cellcolor{blue!47} 94.8\\
\hline
\end{tabular}
}
\caption{[Task 2: Enumeration] Iterative ProofWriter is better at generating all implications than
an All-At-Once strategy. (All-At-Once is trained on D3+Enum,
Iterative ProofWriter is the same model as earlier.) \label{enumeration}}
\vspace{-3mm}
\end{table}

\subsection{Task 3: Abduction (Single Fact)}

Fourth and finally, we evaluate performance on a second new task, namely {\it abduction} over
natural language theories, again made possible by generative models.
Analogous to implication enumeration, we create a derivative abductive dataset
of $C, Q \rightarrow factM_{1}, ..., factM_{i}$ examples, where $C \cup \{factM_{i}\}$
results in $Q$ becoming provable
as described in Section~\ref{abduction-method}. We create such D*-Ab datasets from the D*(OWA) datasets. 

\subsubsection{Results (IID)}

We trained a model on D3-Ab (train), and then tested on both D3-Ab (test) and D5-Ab (test).
We evaluate the results by comparing the predicted
and gold $factM$s, measuring both F1 and ``perfect match'' Accuracy (1 when F1=1, 0 otherwise).
The results are shown in Table~\ref{abduction}, and indicate that the model performs well
overall (85\%+ scores). We also broke down the recall of $factM$s by proof depth required
to prove $Q$ given $C$ and $factM$. This is shown in Table~\ref{abduction-by-depth},
indicating that it is harder to identify a $factM$ that completes a deeper proof.
The similarity of D3-Ab and D5-Ab scores suggests that D5-Ab
is not out-of-domain for this task: Although depths for {\it provable} D5 facts are
deeper than D3, this task concerns {\it unprovable} 
facts, which
may not be distributed differently to D3-Ab.


\begin{table}
\centering
{\small
\setlength{\tabcolsep}{3pt}	
\begin{tabular}{lccc} \hline
{\bf Test:}	& {\bf Count} & {\bf F1}	& 	{\bf Acc} \\ \hline
D3-Ab   & 7067	& \cellcolor{blue!49}97.4	& \cellcolor{blue!47}94.5 \\
D5-Ab   & 7181	& \cellcolor{blue!49}97.3	& \cellcolor{blue!47}93.5 \\ \hline
\end{tabular}
}
\caption{[Task 3: Abduction] 
  Given a theory $C$ and
an unprovable conclusion $Q$, predict all alternative facts that,
when added to $C$, make $Q$ provable. \label{abduction}}
\vspace{-3mm}
\end{table}

\begin{table}
\centering
{\small
\setlength{\tabcolsep}{3pt}	
\begin{tabular}{lcccc} \hline
{\bf Gold Proof} & \multicolumn{2}{c}{\bf Test on D3-Ab} &  \multicolumn{2}{c}{\bf Test on D5-Ab} \\
{\bf  Depth}     & {\bf \# Gold} & {\bf Acc (recall)} & {\bf \# Gold} & {\bf Acc (recall)} \\
\hline
N/A	& 2155	& \cellcolor{blue!49}97.73	& 2170	& \cellcolor{blue!49}97.74 \\
1	& 4813	& \cellcolor{blue!49}98.46	& 4731	& \cellcolor{blue!49}98.73 \\
2	& 1719	& \cellcolor{blue!48}96.22	& 1986	& \cellcolor{blue!48}96.17 \\
3	& 688	& \cellcolor{blue!45}90.26	& 915	& \cellcolor{blue!46}92.79 \\
4	& 153	& \cellcolor{blue!38}75.82	& 330	& \cellcolor{blue!41}82.73 \\
5	& 19	& \cellcolor{blue!18}36.84	& 96	& \cellcolor{blue!39}78.13 \\ \hline
\end{tabular}
}
\caption{[Task 3: Abduction] Recall of abduced facts by proof depth. The data
suggests that it is harder to identify a $factM$ that completes a deeper proof. \label{abduction-by-depth}}
\vspace{-3mm}
\end{table}

\subsubsection{Generalization to New Tasks \label{abduction-ood}}

To assess out-of-domain generalization, we also evaluate how well the trained abductive model
performs on an abductive version of the Birds-Electricity(OWA) theories, zero-shot
(created using the same approach, Section~\ref{abduction-method}). We find that
ProofWriter has perfect zero-shot performance for the simple Birds rulebases, but progressively
reduced performance for the Electricity theories as they get more complex (dropping to 64\% F1, 62\% Accuracy
for one rulebase), indicating that the abductive task is only partly solved 
(Appendix~\ref{abduction-ood-details}).


\section{Discussion}

\subsection{All-At-Once vs. Iterative Strategies}


While the All-At-Once approach to proof generation is simple, efficient, and effective,
it does not generalize as well to proofs of greater depth than seen at training. 
In contrast, the Iterative approach is robust to generalization. Even though
errors at each iteration accumulate, the reliability of 1-step inference is so high
that such error accumulations remain small. The Iterative architecture, namely a simple model embedded in a
recursive loop (rather than single seq2seq model), illustrates how transformers can
be used in a ``scale-invariant''
way,
i.e., performance is largely unchanged by the scale (here reasoning depth) of the problem.
In addition, as proofs are built from actual inference steps taken by the model, they
are by definition ``faithful'' to the model's inference steps, rather than being a post hoc rationalization.

However, there are also some drawbacks to the Iterative approach: First, it is inefficient and
unguided, proving everything possible and only then looking for the answer and proof for a particular question.
In fact, this is a limitation of unconstrained forward-chaining in general, hence
established techniques for guiding forward-chaining could be applied, e.g.,
a best-first expansion strategy, or using a backward-chaining strategy instead
(which would similarly need to be controlled).
Second, as the theory grows by one fact per iteration, there is a risk of exceeding the transformer's input token
limit (512 tokens by default), hence limiting the size of theories that can be handled.
For larger theories, a retrieval mechanism might be needed to manage the facts
and rules available to the reasoner.

\subsection{Abduction and Implicit Knowledge}

Recently, LeapOfThought \cite{leapofthought} showed that RuleTaker-like models could be retrained
to reason with a combination of explicit and implicit knowledge, rather than requiring all rules to be
stated explicitly (the implicit knowledge coming from the latent knowledge acquired during pretraining \cite{Petroni2019LanguageMA}).
Now, given an abductive capability such as the one we have presented, we have a mechanism for
materializing the implicit knowledge used to answer a question, and hence
generating the full proof of its answer: Given a LeapOfThought conclusion, first abduce
the ``missing'' (implicit) fact(s) required for an explicit proof, then use ProofWriter
to generate that proof. This is a significant step forward to help understand
a model's decisions when both implicit and explicit knowledge has been used.

\section{Summary and Conclusion}

While it is remarkable that transformers can learn to systematically reason over language,
such methods will have limited impact if they cannot also explain their answers. In this work,
we showed the first application of generative techniques to this task, and demonstrated how
proofs, implication enumerations, and abductive inferences can be generated,
exceeding the prior state-of-the-art in proof generation by +9\% (absolute).
In addition, the Iterative ProofWriter robustly generalizes to deeper proofs and more
varied language than seen in training, and produces proofs that reflect (i.e., are faithful to)
the model's actual inference decisions. Finally, the abductive capability offers
the potential for generating proofs when both explicit and implicit knowledge are used,
by materializing the implicit knowledge needed to complete the proof.
Together, these significantly improve the viability of 
neural methods for systematically reasoning over language in practical settings.
The ProofWriter datasets are available at https://allenai.org/data/proofwriter

\vspace{2mm}
\noindent 
{\bf Acknowledgements:} We thank Google for providing the TPUs for conducting experiments.

\eat{
Despite this, the presented work is only a first step, working with small theories (subject
to the transformers' token limit) and limited variation in language. Generating proofs
with larger theories may require adding a retrieval component, and greater use
of implicit knowledge will require expanding the abductive capability.
Nevertheless, ProofWriter shows that high-quality, faithful, natural language proofs
can be generated to explain reasoned answers, significantly improving the
viability of systematic reasoning over language for practical applications.
The ProofWriter datasets are available at https://allenai.org/data/proofwriter
}


\bibliography{references}
\bibliographystyle{acl_natbib}

\clearpage

\appendix

\twocolumn[{\centering {\Large {\bf Appendix: ProofWriter: Generating Implications, Proofs, \vspace{2mm} \\ and Abductive Statements over Natural Language \vspace{5mm}}}}]

\section{Datasets: Additional Details}

\subsection{Statistics \label{appendix-statistics}}

Some overall statistics for the updated RuleTaker CWA and OWA datasets are in Table~\ref{dataset-stats}. The number of implications per theory can reach 20 and above, and the proof depths go up to 10, even though the proof depths of the associated questions are limited to the dataset depth (e.g., depth 3 for D3).

\begin{table} [h]
\centering
{\small
\setlength{\tabcolsep}{3pt}	
\begin{tabular}{lccccc}
\hline
 &  &  & \# impl & depth\\
Dataset & \# th & \# qns & min/mean/max & max\\
\hline
{\bf CWA:} &  &  &  & \\
D0 & 27020 & 100002 & 0/1.0/18 & 8\\
D1 & 12965 & 100012 & 1/1.9/17 & 6\\
D2 & 9138 & 100014 & 2/3.3/18 & 5\\
D3 & 7067 & 100024 & 3/5.1/16 & 7\\
D5 & 4935 & 100030 & 5/9.8/21 & 10\\
Birds/Elec & 140 & 5270 & 0/2.0/6 & 4\\
ParaRules & 2403 & 40022 & 3/4.3/14 & 5\\\hline
{\bf OWA:} &  &  &  & \\
D0 & 26978 & 100000 & 0/0.8/18 & 5\\
D1 & 12933 & 100014 & 1/1.7/14 & 6\\
D2 & 9033 & 100010 & 2/3.1/14 & 5\\
D3 & 6940 & 100036 & 3/4.8/16 & 6\\
D5 & 4752 & 100030 & 5/9.1/21 & 10\\
Birds/Elec & 140 & 5270 & 0/1.2/6 & 3\\
ParaRules & 2403 & 40022 & 3/4.3/14 & 5\\
\hline
\end{tabular}
}
\caption{Statistics for the CWA and OWA datasets, giving the number of theories, questions and implications per theory. Note that the maximum implication proof depth can go higher than the maximum proof depth for the included questions (e.g., for D5 the maximum questions depth is 5, but there are implications up to depth 10 which are include in the enumeration task).
\label{dataset-stats}}
\end{table}

\noindent
Table \ref{Ab-dataset-stats} describes overall statistics for the datasets for Task 3: Abduction. Each abduction question can have zero or more missing facts as answer, and the proof depths can go up to 11.
\begin{table} [h]
\centering
{\small
\setlength{\tabcolsep}{3pt}	
\begin{tabular}{lccccc}
\hline
 &  &  & \# missing  & max\\
  &  &  & facts & proof\\
Dataset & \# th & \# qns & min/mean/max & depth\\
\hline
D0-Ab & 18011 & 85705 & 0/0.8/15 & 6 \\
D1-Ab & 10448 & 49808 & 0/0.8/12 & 7 \\
D2-Ab & 7092 & 37245 & 0/0.9/11 & 6 \\
D3-Ab & 5633 & 34915 & 0/1.1/11 & 8 \\
D5-Ab & 4362 & 35213 & 0/1.2/9 & 11 \\
Birds-Electricity-Ab & 140 & 3940 & 0/0.24/4 & 4 \\
\hline
\end{tabular}
}
\caption{Statistics for the Abduction datasets, giving the number of theories, abduction questions, number of missing facts per question and maximum proof depth.
\label{Ab-dataset-stats}}
\end{table}

\subsection{Repairs to the RuleTaker Datasets \label{dataset-repairs}}

The original RuleTaker theories were intended to be full Datalog theories, but contained three occasional violations
in the with-negation theories:
\begin{enu}
\item[1.] Some theories contained negated facts (e.g., ``Bob is not red'') and/or rules with negated conclusions.
Such statements are redundant under a CWA, and not allowed according to formal Datalog specifications.
\item[2.] Some theories included rules with a free variable in a negated condition (e.g., ``If someone is not blue then Bob is happy.'').
Such rules are not allowed according to formal Datalog specifications, as the possible groundings of the variable
require meta-information about the theory as a whole.
\item[3.] A bug in the stratification checker led to a few theories being included that were not stratifiable,
and hence may have multiple, valid truth assignments for their facts.
\end{enu}
As a result, the theories were regenerated (with the same distribution over number of facts, rules, condition, etc.) to create the D*(CWA) datasets,
avoiding these issues.


The D*(OWA) datasets are similar to the D*(orig) datasets, but evaluated without a CWA, i.e., negation-as-failure (NAF) is replaced with hard negation. The theories with negation were again regenerated to ensure they were stratifiable (to avoid negation cycles), but they still retain negated facts and rule conclusions. The truth values of the questions were recomputed using an OWA, resulting in answers {\it True/False/Unknown}. 

\subsection{The Birds-Electricity Datasets \label{dataset-details-birds}}

The RuleTaker ``birds'' rulebase is a well-known logic problem illustrating the use of ``abnormality''
predicates \cite{McCarthy1984ApplicationsOC},\footnote{https://www.doc.ic.ac.uk/$\sim$mjs/teaching/KnowledgeRep491/ ExtendedLP\_491-2x1.pdf, p5},
and converted into English by hand. The dataset contains a single theory of six rules (e.g., ``If someone is a bird and wounded then they are abnormal.'')
and seven facts (e.g., ``Bill is wounded''), and forty questions against this theory (i.e., 40 test examples total). Birds1 and Birds2 differ solely
in the English wording (e.g., ``Bill is flying'' vs. ``Bill can fly'').

The four RuleTaker ``electricity'' datasets contain examples of reasoning about toy electrical cicuits
using a small set of general rules about circuits. Examples in each dataset are built using a fixed
set of general rules per dataset, ranging from five rules (Elec1) to twelve rules (Elec4).
Each example in these datasets contains the general rules, plus between two and five facts describing a
particular circuit, with a set of questions about the circuit, e.g., Q: ``The light bulb is glowing?'' A: {\it True}.

\subsection{The ParaRules Dataset \label{dataset-details-para}}

The RuleTaker ``ParaRules'' dataset contains 40k questions against 2k theories expressed in paraphrased natural language, obtained
by having crowdworkers rephrase the templated English facts and rules from sampled original theories into more varied natural language.
For example, ``Bob is cold.'' might be rephrased ``In the snow sits Bob, crying from being cold''; or
``Alan is round. Alan is blue. Alan is kind.'' might be rephrased ``Alan, who is round and also kind, tends to be rather blue'';
or ``If someone is kind then they are young.'' might be rephrased ``A kind person will certainly be young.''.
While the previous datasets contain synthetic language, ParaRules tests the models’ ability to reason over more human-like paraphrased
language.

\section{Additional Results}

\subsection{Results on the OOD ParaRules Dataset \label{appendix-pararules}}

\begin{table}
\centering
{\small
\setlength{\tabcolsep}{3pt}	
\begin{tabular}{llcccc} \hline
 &   & \multicolumn{2}{c}{\bf Answer} & \multicolumn{2}{c}{\bf Proof} \\
      &	\# qns	            &   PRover & ProofWriter & PRover & ProofWriter \\ 
\hline
D=0 & 2968 & \cellcolor{blue!50} 99.7 & \cellcolor{blue!50} 99.9 & \
\cellcolor{blue!50} 99.4 & \cellcolor{blue!50} 99.9\\
1 & 2406 & \cellcolor{blue!49} 98.6 & \cellcolor{blue!50} 99.3 & \
\cellcolor{blue!49} 97.3 & \cellcolor{blue!50} 99.3\\
2 & 1443 & \cellcolor{blue!49} 98.2 & \cellcolor{blue!49} 98.3 & \
\cellcolor{blue!44} 88.7 & \cellcolor{blue!49} 97.7\\
3 & 1036 & \cellcolor{blue!48} 96.5 & \cellcolor{blue!49} 98.2 & \
\cellcolor{blue!45} 89.9 & \cellcolor{blue!48} 96.5\\
4 & 142 & \cellcolor{blue!44} 88.0 & \cellcolor{blue!46} 91.5 & \
\cellcolor{blue!38} 76.1 & \cellcolor{blue!42} 83.1\\\hline
\bf{All} & 8008 & \cellcolor{blue!49} 98.4 & \cellcolor{blue!50} 99.1 & \
\cellcolor{blue!48} 95.1 & \cellcolor{blue!49} 98.5\\
\hline      
\end{tabular}
}
\caption{[Task 1: Proof Generation] Train on D3 + ParaRules, test on (only) ParaRules. Both systems
demonstrate robustness to more complex linguistic expressions
in the theories, with ProofWriter obtaining 3\% higher proof correctness. \label{pararules}}
\end{table}

We also test the robustness of ProofWriter's proof generation to theories that use more varied natural
language, summarized in Section~\ref{proofs-on-ood}. Following \cite{ruletaker} and \cite{prover}, we train on the combined training partitions of D3(orig) and ParaRules,
then test on the ParaRules test partition. The results in Table~\ref{pararules} show that
PRover and ProofWriter (All-At-Once) are robust to more complex natural language
in the input, with ProofWriter obtaining 3\% higher proof correctness.

\subsection{Abduction: Generalization to New Tasks \label{abduction-ood-details}}

Section~\ref{abduction-ood} summarized the results of testing abductive reasoning
on abductive versions of the Birds-Electricity(OWA) theories. The detailed results
are shown in Table~\ref{ood-abduction}, showing perfect zero-shot performance for the simple Birds rulebases, but progressively
reduced performance for the Electricity theories as they get more complex.
This indicates that the abductive task remains only partially solved by our generative model.

\begin{table}
\centering
{\small
\setlength{\tabcolsep}{3pt}	
\begin{tabular}{lccc} \hline
{\bf Test Dataset:}     & {\bf \# qns} & {\bf F1} & {\bf Acc} \\ \hline
Birds1-Ab & 14 & \cellcolor{blue!50}100.00 & \cellcolor{blue!50}100.00 \\
Birds2-Ab & 14 & \cellcolor{blue!50}100.00 & \cellcolor{blue!50}100.00 \\
Elec1-Ab & 114 & \cellcolor{blue!45}89.47 & \cellcolor{blue!45}89.47 \\
Elec2-Ab & 126 & \cellcolor{blue!45}90.25 & \cellcolor{blue!44}88.89 \\
Elec3-Ab & 456 & \cellcolor{blue!41}81.79 & \cellcolor{blue!38}76.32 \\
Elec4-Ab & 3216 & \cellcolor{blue!43}85.77 & \cellcolor{blue!42}83.99 \\\hline
All & 3940 & \cellcolor{blue!43}85.66 & \cellcolor{blue!42}83.53 \\ \hline
\end{tabular}
}
\caption{[Task 3: Abduction] Zero-shot scores of the D3-Ab model on the Birds-Electricity-Ab rulebases.  \label{ood-abduction}}
\end{table}

\section{Results with T5-large \label{appendix-t5-large}}

In the main part of the paper we trained ProofWriter starting from the largest available T5-11B model (11 billion parameters). If we instead use the more manageable T5-large model (770 million parameters), the scores generally go down, but typically by a small amount. 

In Tables~\ref{compare-large-11b-d3-on-d5} and \ref{compare-large-11b-d3iter-on-d5} we show two examples of this, for the All-At-Once and Iterative ProofWriter models respectively, when training on the D3 dataset and evaluating on D5. We see the T5-large model is a bit worse on higher depth proof accuracy in the All-At-Once model, but is otherwise quite competitive. 

\begin{table}
\centering
{\small
\setlength{\tabcolsep}{3pt}	
\begin{tabular}{lccccccccc}
\hline
 & \multicolumn{4}{c}{\bf{Answer}} & \multicolumn{4}{c}{\bf{Proof}}\\
 & \multicolumn{2}{c}{CWA} & \multicolumn{2}{c}{OWA} & \multicolumn{2}{c}{CWA} & \multicolumn{2}{c}{OWA}\\
Depth & large & 11B & large & 11B & large & 11B & large & 11B\\
\hline
N/A & \cellcolor{blue!49} 98.4 & \cellcolor{blue!50} 99.6 & \cellcolor{blue!49} 97.4 & \cellcolor{blue!50} 99.4 & \cellcolor{blue!49} 98.4 & \cellcolor{blue!50} 99.6 & \cellcolor{blue!49} 97.4 & \cellcolor{blue!50} 99.4\\
0 & \cellcolor{blue!50} 100 & \cellcolor{blue!50} 100 & \cellcolor{blue!50} 100 & \cellcolor{blue!50} 100 & \cellcolor{blue!50} 100 & \cellcolor{blue!50} 100 & \cellcolor{blue!50} 100 & \cellcolor{blue!50} 100\\
1 & \cellcolor{blue!50} 100 & \cellcolor{blue!50} 99.9 & \cellcolor{blue!50} 99.9 & \cellcolor{blue!50} 99.9 & \cellcolor{blue!50} 99.4 & \cellcolor{blue!50} 99.7 & \cellcolor{blue!50} 99.3 & \cellcolor{blue!50} 99.8\\
2 & \cellcolor{blue!50} 99.8 & \cellcolor{blue!50} 99.4 & \cellcolor{blue!50} 99.7 & \cellcolor{blue!50} 99.8 & \cellcolor{blue!49} 97.5 & \cellcolor{blue!49} 98.2 & \cellcolor{blue!49} 97.6 & \cellcolor{blue!49} 98.8\\
3 & \cellcolor{blue!50} 100 & \cellcolor{blue!50} 99.2 & \cellcolor{blue!50} 99.7 & \cellcolor{blue!50} 99.8 & \cellcolor{blue!45} 90.4 & \cellcolor{blue!47} 93.4 & \cellcolor{blue!46} 91.2 & \cellcolor{blue!47} 94.5\\
4 & \cellcolor{blue!49} 98.9 & \cellcolor{blue!48} 95.4 & \cellcolor{blue!50} 99.5 & \cellcolor{blue!50} 99.3 & \cellcolor{blue!19} 38.6 & \cellcolor{blue!35} 69.9 & \cellcolor{blue!23} 46.9 & \cellcolor{blue!36} 71.4\\
5 & \cellcolor{blue!46} 92.3 & \cellcolor{blue!36} 72.9 & \cellcolor{blue!49} 98.9 & \cellcolor{blue!47} 93.7 & \cellcolor{blue!6} 12.4 & \cellcolor{blue!14} 27.4 & \cellcolor{blue!12} 24.4 & \cellcolor{blue!18} 35.1\\\hline
All & \cellcolor{blue!49} 98.4 & \cellcolor{blue!48} 96.6 & \cellcolor{blue!49} 98.7 & \cellcolor{blue!50} 99.0 & \cellcolor{blue!42} 83.4 & \cellcolor{blue!44} 88.9 & \cellcolor{blue!43} 85.6 & \cellcolor{blue!45} 90.2\\
\hline
\end{tabular}
}
<\caption{[Task 1] Comparing T5-large vs T5-11B for the All-At-Once models trained on D3 and evaluated on D5. T5-large is actually a bit ahead of T5-11B on answer accuracy (for CWA), although the proof correctness is noticeably higher with T5-11B.
  \label{compare-large-11b-d3-on-d5}}
\vspace{-2mm}
\end{table}

\begin{table}
\centering
{\small
\setlength{\tabcolsep}{3pt}	
\begin{tabular}{lccccccccc}
\hline
 & \multicolumn{4}{c}{\bf{Answer}} & \multicolumn{4}{c}{\bf{Proof}}\\
 & \multicolumn{2}{c}{CWA} & \multicolumn{2}{c}{OWA} & \multicolumn{2}{c}{CWA} & \multicolumn{2}{c}{OWA}\\
Depth & large & 11B & large & 11B & large & 11B & large & 11B\\
\hline
N/A & \cellcolor{blue!50} 99.0 & \cellcolor{blue!50} 99.7 & \cellcolor{blue!50} 99.2 & \cellcolor{blue!50} 99.9 & \cellcolor{blue!50} 99.0 & \cellcolor{blue!50} 99.7 & \cellcolor{blue!50} 99.2 & \cellcolor{blue!50} 99.9\\
0 & \cellcolor{blue!50} 100 & \cellcolor{blue!50} 100 & \cellcolor{blue!50} 100 & \cellcolor{blue!50} 100 & \cellcolor{blue!50} 100 & \cellcolor{blue!50} 100 & \cellcolor{blue!50} 100 & \cellcolor{blue!50} 100\\
1 & \cellcolor{blue!49} 98.8 & \cellcolor{blue!50} 99.8 & \cellcolor{blue!50} 99.1 & \cellcolor{blue!50} 99.3 & \cellcolor{blue!48} 95.0 & \cellcolor{blue!48} 95.4 & \cellcolor{blue!49} 97.5 & \cellcolor{blue!49} 97.8\\
2 & \cellcolor{blue!49} 98.3 & \cellcolor{blue!50} 99.5 & \cellcolor{blue!49} 98.9 & \cellcolor{blue!50} 99.7 & \cellcolor{blue!46} 91.0 & \cellcolor{blue!46} 91.7 & \cellcolor{blue!48} 96.4 & \cellcolor{blue!49} 97.3\\
3 & \cellcolor{blue!49} 98.6 & \cellcolor{blue!50} 99.7 & \cellcolor{blue!49} 98.4 & \cellcolor{blue!50} 99.2 & \cellcolor{blue!44} 89.0 & \cellcolor{blue!45} 90.4 & \cellcolor{blue!48} 95.5 & \cellcolor{blue!49} 97.1\\
4 & \cellcolor{blue!49} 98.0 & \cellcolor{blue!50} 99.7 & \cellcolor{blue!49} 97.5 & \cellcolor{blue!50} 99.1 & \cellcolor{blue!43} 86.3 & \cellcolor{blue!44} 88.9 & \cellcolor{blue!47} 93.4 & \cellcolor{blue!48} 96.5\\
5 & \cellcolor{blue!49} 97.7 & \cellcolor{blue!49} 98.9 & \cellcolor{blue!48} 96.5 & \cellcolor{blue!49} 98.8 & \cellcolor{blue!43} 85.4 & \cellcolor{blue!44} 87.8 & \cellcolor{blue!41} 82.3 & \cellcolor{blue!43} 86.4\\\hline
All & \cellcolor{blue!49} 98.7 & \cellcolor{blue!50} 99.7 & \cellcolor{blue!49} 98.8 & \cellcolor{blue!50} 99.6 & \cellcolor{blue!47} 94.4 & \cellcolor{blue!48} 95.4 & \cellcolor{blue!48} 96.4 & \cellcolor{blue!49} 97.6\\
\hline
\end{tabular}
}
\caption{[Task 1] Comparing T5-large vs T5-11B for the Iterative ProofWriter trained on D0-D3 and evaluated on D5. T5-11B is systematically slightly better.
  \label{compare-large-11b-d3iter-on-d5}}
\vspace{-2mm}
\end{table}


\section{Hyperparameters and I/O Examples \label{appendix-inputs-outputs}}

We fine-tune the models on the training set using the default hyperparameters (including the Adafactor optimizer) in the T5 library.\footnote{https://github.com/google-research/text-to-text-transfer-transformer} We use the largest T5-11B model for the main results, fine-tuned for 40k steps (batch size 8), selecting the checkpoint with highest validation score (usually the final step).  See Appendix~\ref{appendix-t5-large} for results using the smaller T5-large. 

For all tasks in this paper, the model input and output are plain textual strings in a certain format. The format is slightly more verbose than strictly necessary, as it is compatible with a more general framework combining different views of the task. This is not typically an issue, although there are a few cases where the iterative model reaches the default 512 token limit of T5 when adding implications to the theory.
We now show input/output examples for proof generation (All-At-Once and (one step of) the Iterative approaches), implication generation (All-At-Once approach; Iterative implication generation is a side-effect of Iterative proof generation), and abduction.

\subsection{All-At-Once Proof Generation \label{appendix-all-at-once-input-output}}

In this task we are given a theory and question, and need to produce an answer (True/False/Unknown) and proof with intermediates.

\vspace{1mm}
\begin{spacing}{0.8}
{\small {
\noindent{\bf INPUT:} \$answer\$ ; \$proof\$ ; \$question\$ = The lion is not nice? ; \$context\$ = sent1: The tiger chases the lion. sent2: The lion is not big. sent3: If something visits the dog and it is kind then it visits the mouse. sent4: The dog chases the lion. sent5: If something is big then it visits the dog. sent6: The tiger eats the dog. sent7: If something visits the tiger and the tiger is not red then it is not nice. sent8: If something chases the dog then it is not red. sent9: The mouse does not chase the tiger. sent10: If something visits the mouse then the mouse is red. sent11: The tiger visits the lion. sent12: The tiger does not eat the mouse. sent13: The mouse is nice. sent14: The lion does not eat the tiger. sent15: If the tiger visits the dog and the dog is not big then the dog chases the mouse. sent16: The lion visits the tiger. sent17: If something chases the lion and it visits the dog then it chases the dog. sent18: The dog is nice. sent19: If something chases the lion then it is big. sent20: If something eats the lion then it is not cold.\\
{\bf OUTPUT:} \$answer\$ = True ; \$proof\$ = \# sent7@int1 \& sent16 \# sent8@int2 \# sent17@int3 \& sent1 \# sent5@int4 \# sent19@int5 sent1 ; with int1: The lion is not nice. ; int2: The tiger is not red. ; int3: The tiger chases the dog. ; int4: The tiger visits the dog. ; int5: The tiger is big.
}}
\end{spacing}
\vspace{1mm}

If we translate back from Polish Notation, this proof corresponds to ((sent16 \& ((((sent1 \&  ((sent1 $\rightarrow$ sent19 $\rightarrow$ int5) $\rightarrow$ sent5 $\rightarrow$ int4)) $\rightarrow$ sent17 $\rightarrow$ int3)) $\rightarrow$ sent8 $\rightarrow$ int2)) $\rightarrow$ sent7 $\rightarrow$ int1)

\subsection{Iterative Proof Generation (one step) \label{appendix-iterative-input-output}}

In this task we ask the model to generate one valid implication of the theory with proof of depth 1. If no such implication exists, instead generate "None" as the answer.

\vspace{1mm}
\begin{spacing}{0.8}
{\small {
\noindent{\bf INPUT:} \$answer\$ ; \$proof\$ ; \$question\$ = What is one single-hop inference? ; \$context\$ = sent1: If something eats the cow and it is big then the cow sees the bald eagle. sent2: If something likes the bald eagle then it is rough. sent3: If something eats the dog then it likes the cow. sent4: Big things are young. sent5: If something likes the cow then it eats the cow. sent6: If something sees the bald eagle then the bald eagle eats the cow. sent7: If something likes the bald eagle then the bald eagle is kind. sent8: If something sees the bald eagle then the bald eagle eats the dog. sent9: The bald eagle eats the cow. sent10: The bald eagle sees the dog. sent11: The dog is big. sent12: The cow likes the bald eagle. sent13: The bald eagle is young. sent14: The dog sees the cow. sent15: The bald eagle is kind. sent16: The dog is young. sent17: The bald eagle sees the cow. sent18: The bald eagle is rough. sent19: The cow eats the bald eagle. sent20: The dog is cold. sent21: The dog likes the cow. sent22: The dog eats the bald eagle. sent23: The dog eats the cow. sent24: The bald eagle likes the dog. sent25: The bald eagle likes the cow. sent26: The cow sees the bald eagle. sent27: The cow sees the dog.\\
{\bf OUTPUT:} \$answer\$ = The cow is rough. ; \$proof\$ = \# sent2 sent12
}}
\end{spacing}
\vspace{1mm}

During evaluation time, we would then take this output and append it to the theory as "sent28: The cow is rough." and repeat the question. If a subsequent proof then references "sent28", we know already that this was derived using the proof "\# sent2 sent12". 

\subsection{All-At-Once Implication Enumeration \label{appendix-enumeration-input-output}}

In this task we ask the model to simply output all valid implications of a theory, without generations proofs in this case.

\vspace{1mm}
\begin{spacing}{0.8}
{\small  {
\noindent{\bf INPUT:} \$answer\$ ; \$proof\$ ; \$question\$ = What are all the inferences? ; \$context\$ = sent1: The tiger chases the lion. sent2: The lion is not big. sent3: If something visits the dog and it is kind then it visits the mouse. sent4: The dog chases the lion. sent5: If something is big then it visits the dog. sent6: The tiger eats the dog. sent7: If something visits the tiger and the tiger is not red then it is not nice. sent8: If something chases the dog then it is not red. sent9: The mouse does not chase the tiger. sent10: If something visits the mouse then the mouse is red. sent11: The tiger visits the lion. sent12: The tiger does not eat the mouse. sent13: The mouse is nice. sent14: The lion does not eat the tiger. sent15: If the tiger visits the dog and the dog is not big then the dog chases the mouse. sent16: The lion visits the tiger. sent17: If something chases the lion and it visits the dog then it chases the dog. sent18: The dog is nice. sent19: If something chases the lion then it is big. sent20: If something eats the lion then it is not cold.\\
{\bf OUTPUT:} \$answer\$ = The dog is big. The tiger is big. The dog visits the dog. The tiger visits the dog. The dog chases the dog. The tiger chases the dog. The dog is not red. The tiger is not red. The lion is not nice.
}}
\end{spacing}
\vspace{1mm}

This particular theory has 9 valid implications, around average for D5 theories. (The number of implications can be as high as 21).

\subsection{Iterative Implication Enumeration}
\vspace{-1mm}

Iterative implication generation is a side-effect of Iterative proof generation, hence the input/output (one step) is the same as in Appendix~\ref{appendix-iterative-input-output}.

\subsection{Abduction \label{appendix-abduction-input-output}}
\vspace{-1mm}

In this task we ask the model to output all alternative missing facts, separated by a comma, that can be (separately) added to a theory so as to prove the given implication to be True. 
\vspace{1mm}
\begin{spacing}{0.8}
{\small {
\noindent{\bf INPUT:} \$answer\$ ; \$question\$ = Dave is rough. ; \$context\$ = triple1: Anne is white. triple2: Charlie is young. triple3: Dave is round. triple4: Erin is quiet. rule1: If someone is rough and young then they are blue. rule2: Rough, white people are smart. rule3: All smart people are rough. rule4: All white people are smart. rule5: If someone is young then they are smart. rule6: All smart people are rough.\\
{\bf OUTPUT:} \$answer\$ = Dave is young. , Dave is smart.
}}
\end{spacing}
\vspace{1mm}

This particular abduction question has 2 alternative missing facts. Any one of them when added to the input theory is sufficient to prove the implication `Dave is rough'.

\end{document}